%% file: PaperForReview.tex
\documentclass[10pt,twocolumn,letterpaper]{article}

\usepackage{cvpr}              

\usepackage{graphicx}
\usepackage{amsmath}
\usepackage{amssymb}
\usepackage{booktabs}

\usepackage{wrapfig}

\usepackage[dvipsnames]{xcolor}
\usepackage{tabularx}

\usepackage{color}
\usepackage{colortbl}

\newcommand{\drop}[1]{\textcolor{gray}{\textsubscript{--#1}}}
\newcommand{\rise}[1]{\textcolor{gray}{\textsubscript{+#1}}}

\newcommand{\Rise}[1]{\textcolor{Green}{\textsubscript{\bf +#1}}}
\newcommand{\RiseB}[1]{\textcolor{Blue}{\textsubscript{\bf +#1}}}
\newcommand{\band}{\rowcolor{gray!20}}

\usepackage[pagebackref,breaklinks,colorlinks]{hyperref}

\usepackage[capitalize]{cleveref}
\crefname{section}{Sec.}{Secs.}
\Crefname{section}{Section}{Sections}
\Crefname{table}{Table}{Tables}
\crefname{table}{Tab.}{Tabs.}

\def\cvprPaperID{14} 
\def\confName{CVPR}
\def\confYear{2023}

\begin{document}

\title{Estimating and Maximizing Mutual Information for Knowledge Distillation}

\author{
Aman Shrivastava\\
University of Virginia\\
Charlottesville, VA, USA\\
{\tt\small as3ek@virginia.edu}
\and
Yanjun Qi\\
University of Virginia\\
Charlottesville, VA, USA\\
{\tt\small yq2h@virginia.edu}
\and
Vicente Ordonez\\
Rice University\\
Houston, TX, USA\\
{\tt\small vicenteor@rice.edu}}

\maketitle

\begin{abstract}
\input{src/000abstract}
\end{abstract}

\section{Introduction}
\label{sec:introduction}
\input{src/010introduction}

\section{Related Work}
\label{sec:background}
\input{src/020related_works}

\section{Method}
\label{sec:method}
\input{src/030method}

\section{Experiments}
\label{sec:experiments}
\input{src/040experiments}

\section{Conclusion}
\label{sec:discussion}
\input{src/050discussion}

{\small
\bibliographystyle{ieee_fullname}
\bibliography{egbib}
}

\clearpage
\newpage
\appendix
\label{sec:appendix}
\input{src/090appendix}

\end{document}

%% file: src/000abstract.tex
In this work, we propose Mutual Information Maximization Knowledge Distillation (MIMKD). Our method uses a contrastive objective to simultaneously estimate and maximize a lower bound on the mutual information of local and global feature representations between a teacher and a student network. We demonstrate through extensive experiments that this can be used to improve the performance of low capacity models by transferring knowledge from more performant but computationally expensive models. This can be used to produce better models that can be run on devices with low computational resources. Our method is flexible, we can distill knowledge from teachers with arbitrary network architectures to arbitrary student networks. Our empirical results show that MIMKD outperforms competing approaches across a wide range of student-teacher pairs with different capacities, with different architectures, and when student networks are with extremely low capacity. We are able to obtain $74.55\%$ accuracy on CIFAR100 with a ShufflenetV2 from a baseline accuracy of $69.8\%$ by distilling knowledge from ResNet-50. On Imagenet we improve a ResNet-18 network from 68.88\% to 70.32\% accuracy (1.44\%+) using a ResNet-34 teacher network.

%% file: src/010introduction.tex
Recent machine learning literature has seen a lot of progress driven by deep neural networks. Many such models that achieve state-of-the-art performance on different benchmarks require large amounts of computation and memory capacities~\cite{huang2018data}. These requirements limit the wider adoption of these models in resource-limited scenarios. To this end, Knowledge Distillation (KD) has been used to transfer knowledge from a stronger teacher network to a smaller and less computationally expensive student network~\cite{bucilu2006model,hinton2015distilling}. This often allows student networks to outperform identical models trained without a teacher. However there is still much room for improvement in knowledge distillation so that students can extract as much knowledge as possible from the teacher network.

\begin{figure}[t]
\centering
\includegraphics[width=0.99\linewidth]{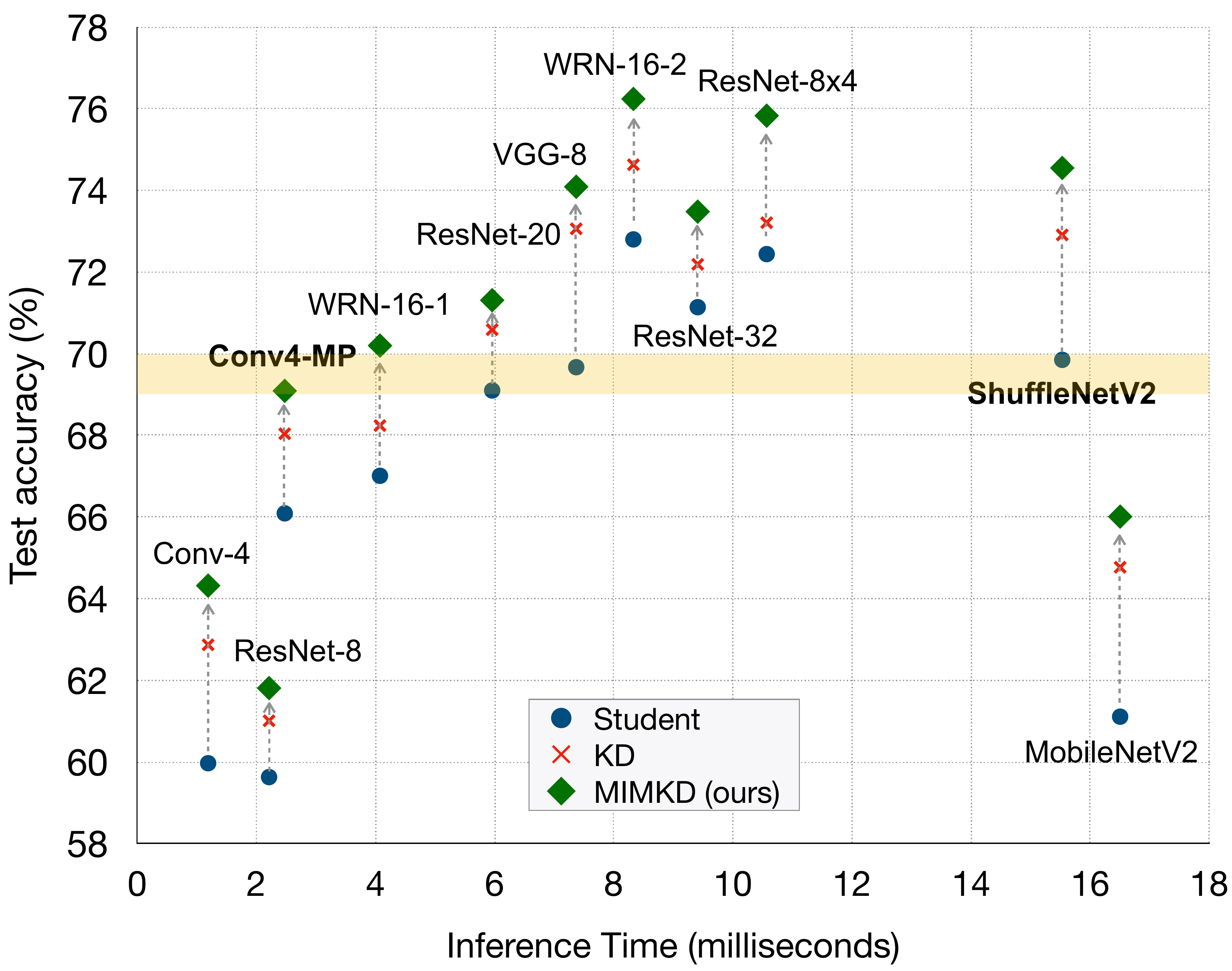}
   \caption{Accuracy-efficiency trade-off of various CNN models on the CIFAR100 dataset. Inference time is the average over a single input. MIMKD provides significant gains over baseline student network accuracies and KD~\cite{hinton2015distilling} across a range of student-teacher pairs. 
   Notice that Conv4-MP, which is a simple 4-layer CNN, after distillation performs close to a more standard ShuffleNetV2.
   \textit{Note:} Runtimes were computed on an Intel(R) Xeon(R) Gold 5218 CPU @ 2.30GHz.
   }
\label{fig:long}
\label{fig:onecol}
\end{figure}

In this paper, we look at knowledge distillation from an information-theoretic perspective. For better distillation, the student needs to generate representations that share maximum information with the representations generated by the teacher. Based on this intuition, we propose Mutual Information Maximization Knowledge Distillation (MIMKD). Multiple approaches have been proposed to estimate the mutual information between high-dimensional continuous variables~\cite{belghazi2018mutual,hjelm2018learning}. Belghazi~et~al~\cite{belghazi2018mutual} propose a KL-divergence based formulation of mutual information. We observe that this approach can be extended to maximize the mutual information in a contrastive setup. Contrastive methods have had an outsized impact in other problems such as self-supervised learning~\cite{chen2020simple,he2020momentum}, however they rely on sampling a rather large number of paired inputs to optimize their objective functions. We find that by using a Jensen-Shannon divergence (JSD) based formulation we obtain a more stable objective to optimize where the performance is invariant to the number of negative samples while being monotonically related to the true mutual information as also shown in Hjelm~et~al~\cite{hjelm2018learning}.

The previous work of~\cite{hinton2015distilling} performed distillation by minimizing the Kullback–Leibler Divergence (KLD) between the output logits of two models. Since then, several other output-based knowledge distillation approaches have been developed~\cite{hinton2015distilling,zhang2019fast,chen2017learning,tian2019contrastive}. Some of these methods try to match the final outputs of two networks by minimizing a distance metric. Other works have also encouraged additional knowledge transfer by minimizing a metric between intermediate representations~\cite{romero2014fitnets,zagoruyko2016paying,ahn2019variational}. 

However, models with significantly different architectures have distinct data-abstraction flows, and the complexity of the patterns recognized at different depths in the model varies significantly with model architecture (i.e.~the number of filters in convolutional layers). Therefore, minimizing a non-parameterized distance metric on the representations imposes an additional structural constraint that might not be ideal for knowledge transfer. In our work, we are still able leverage both local and global information by maximizing mutual information instead of a rigid distance metric between representations. 

More comparable to our work is the recently proposed Contrastive Representation Distillation (CRD) framework~\cite{tian2019contrastive}. This method uses a Noise Contrastive Estimation (NCE) objective~\cite{oord2018representation,gutmann2010noise} to transfer structured relational knowledge from the teacher to the student. However, a caveat of this approach is that it ignores intermediate distillation for feature level information and requires a large number of negative samples requiring large batches~\cite{chen2020simple} or memory banks~\cite{he2020momentum,wu2018unsupervised}. We extend this work by using a JSD-based contrastive objective that is insensitive to the number of negative samples. This enables us to impose additional region-consistent local and feature-level constraints with just one negative sample. 

We propose three mutual information maximization objectives between the teacher and student networks: (1) \emph{Global information maximization}, which aims to maximize the shared information between the final output representations. This pushes the student network to generate feature vectors that are as rich as the ones generated by the teacher. (2) \emph{Local information maximization}, which pushes the student network to recognize complex patterns from each region of the image that are ultimately useful for classification. This is achieved by maximizing the mutual information between region-specific vectors extracted from an intermediate representation of the student network and the final representation of the teacher network. Finally, (3) \emph{Feature Information Maximization}, which is designed to structurally improve the granular feature-extraction capability of the student by maximizing the mutual information between region-consistent local vectors extracted from intermediate representations of the networks. 

Our experimental results in Section~\ref{sec:experiments} demonstrate that these objectives are effective across a wide range of student-teacher pairs and conduct extensive ablation studies of the effect of each proposed objective. We particularly demonstrate the effectiveness of our method in knowledge distillation across student-teacher network pairs with different capacities, student-teacher network pairs with different architectures, and in the extreme case where student networks are extremely low capacity. We show that MIMKD provides consistently better results across all these testing scenarios. Figure~\ref{fig:long} shows a summary of results for various student networks on CIFAR-100 when compared to regular KLD-based knowledge distillation~\cite{hinton2015distilling}. Moreover, we compare the transferability of features learned with knowledge distillation from MIMKD and SOTA baseline. Our results show that MIMKD learns general and transferable representations. 

\begin{figure*}[h]
\begin{center}
\includegraphics[width=\linewidth]{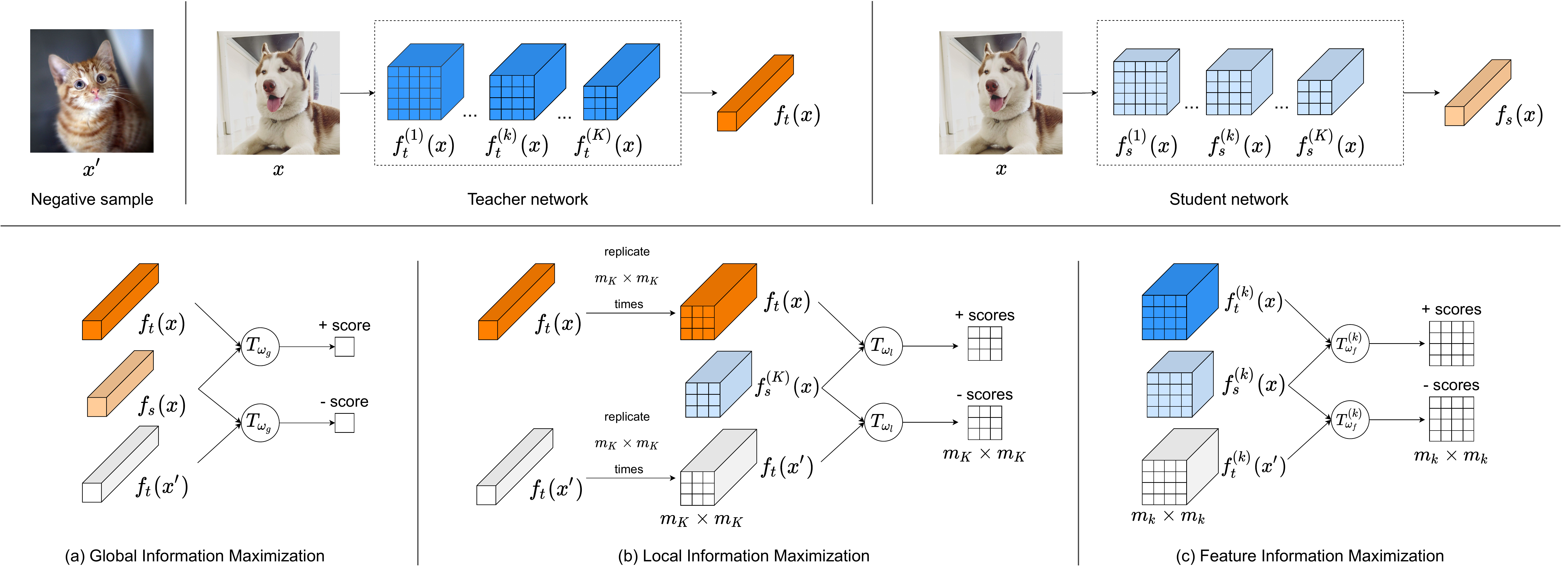}
\end{center}
   \caption{Overall schematic of our proposed method for mutual information maximization based knowledge distillation (MIMKD). \textit{\textbf{Top:}} Representations generated by teacher and student networks for image $x$ and a negative sample $x'$. Note that our method uses only one negative sample. \textbf{\textit{Bottom:}} \textbf{(a)} Positive and negative pairs of final feature vectors are passed into the discriminator function to get scores. \textbf{(b)} Teacher's final representation is replicated to match student's last intermediate representation. \textbf{(c)} For each group of same-sized intermediate feature maps in set $\mathcal{R}$, positive and negative pairs are passed into a distinct discriminator function to get scores. The positive and negative scores obtained are then used with equation (2) to estimate and maximize a lower-bound on mutual information.}
\label{fig:mi_types}
\end{figure*}

%% file: src/020related_works.tex
In this section, we discuss previous efforts in improving knowledge distillation, and in estimating mutual information which are the key areas of contribution of our work. 

\subsection{Knowledge distillation.} The concept of knowledge distillation (KD) was introduced in the works of Buciluǎ~et al.~\cite{bucilu2006model} and later formalized for deep neural networks by Hinton~et al.~\cite{hinton2015distilling}. In knowledge distillation, the goal is to train smaller models that can mimic the performance of larger models. Hinton~et al.~\cite{hinton2015distilling} proposed a knowledge distillation method in which the student network is trained using soft labels extracted from teacher networks. %

Attention transfer~\cite{zagoruyko2016paying} introduced the idea of transferring intermediate attention maps from the teacher to the student network. Fitnets~\cite{romero2014fitnets} also presented the idea of adding more supervision by matching the intermediate representation using regressors. Yim~et al.~\cite{yim2017gift} formulated the distillation problem using the flow of solution procedure (FSP), which is computed as the gram matrix of features across layers. Sau~et al.~\cite{sau2016deep} proposed to include a noise-based regularizer while training the student with the teacher. Specifically, they perform perturbation in the logits of the teacher as a regularization approach. In Correlation Congruence for Knowledge Distillation (CCKD)~\cite{peng2019correlation}, the authors present a framework which transfers not only instance-level information but also the correlation between instances. In CCKD, a Taylor series expansion-based kernel method is proposed to better capture the correlation between instances. Tung~et al.~\cite{tung2019similarity} propose a loss that is based on the observation that semantically similar inputs produce similar activation patterns in trained networks. Variational Information Distillation (VID)~\cite{ahn2019variational} uses a variational lower-bound for the mutual information between the teacher and the student representations by approximating an intractable conditional distribution using a pre-defined variational distribution.

More closely related to our work are methods that cast knowledge distillation as a mutual information maximization problem. Contrastive representation distillation (CRD)~\cite{tian2019contrastive} used a contrastive objective similar to Oord~et al.~\cite{oord2018representation} to maximize a lower-bound on mutual information between final representations. The objective used by CRD is a strong lower-bound on the mutual information but requires a significant number of negative samples during training, consequently, requiring large batch-sizes or memory buffers. These practical constraints become even more limiting if mutual information needs to be minimized at the feature-level to enforce regional-supervision during student training. Our work proposes an alternative that bypasses the needed for such large batch-sizes and thus enables to optimize for mutual information through three separate objectives.

\subsection{Mutual Information Estimation.} Mutual information is a fundamental quantity that measures the relationship between random variables but it is notoriously difficult to measure~\cite{paninski2003estimation}. An exact estimate is only tractable for discrete variables or a small set of problems where the probability distributions are know. However, both the mentioned scenarios are unlikely for real-world visual datasets. Recently, Mutual Information Neural Estimation (MINE)~\cite{belghazi2018mutual} demonstrated a strong method for estimation of mutual information between high-dimensional continuous random variables using neural networks and gradient descent. MINE~\cite{belghazi2018mutual} proposed a general-purpose parametric neural estimator of mutual information based on dual representations of the KL-divergence~\cite{ruderman2012tighter}. Following from MINE~\cite{belghazi2018mutual}, Deep InfoMax~\cite{hjelm2018learning} proposed a mutual information based objective for unsupervised representation learning. Deep InfoMax~\cite{hjelm2018learning} contends that it is unnecessary to use the exact KL-divergence based formulation of mutual information and demonstrated the use of an alternative formulation based on the Jensen-Shannon divergence (JSD). The authors showed that the JSD based estimator is stable, and does not require a large number of negative samples. In addition, Deep InfoMax~\cite{hjelm2018learning} also demonstrated the value of including global and local structure-based mutual information objectives for representation learning. We leverage this line of work in our method to propose a framework for knowledge distillation that leverages both local and global features without significantly adding memory overheads during training.

%% file: src/030method.tex
In this section, we describe our general framework for model compression or knowledge distillation in a teacher student setup. Consider a stronger teacher network $f_t: X \to Y$ with trained parameters $\phi$ and a student network, operating on the same domain, $f_s: X \to Y$ with parameters $\theta$. Let $x$ be the sample drawn from the data distribution $p(x)$ and $f_t(x)$ \& $f_s(x)$ denote the representations extracted from the pre-classification layer, while $f^{cls}_t(x)$ \& $f^{cls}_s(x)$ denote the predicted class-probability distributions from the teacher and the student networks respectively. Now consider a set $\mathcal{R} = \{(f_t^{(k)}(x), f_s^{(k)}(x))\}_{k=1}^{K}$ that contains $K$ pairs of intermediate representations extracted from the networks such that each pair in set $\mathcal{R}$ contains same-sized intermediate representations extracted from the networks, where $m_k \times m_k$ is the size corresponding to the $k$-th pair in the set. Each location in these 2-dimensional intermediate representations corresponds to a specific region in the input image. Note that we do not include the final representations $f_t(x)$ and $f_s(x)$ in the set $\mathcal{R}$. 

Our method focuses on maximizing the mutual information, (1) between final image representations $f_t(x)$ and $f_s(x)$ (global information maximization), (2) between the global image representation from the teacher network $f_t(x)$ and the last intermediate representation from the student network $f_s^{(K)}(x)$ (local information maximization), and (3) between the pairs in set $\mathcal{R}$ (feature information maximization). Figure~\ref{fig:mi_types} shows an overview of our method.

\subsection{Mutual Information Maximization}
In order to estimate and maximize mutual information between random variables $X$ and $Z$, we train a neural network to distinguish samples generated from the joint distribution, $P(X, Z)$ and the product of marginals $P(X)P(Z)$. In MINE~\cite{belghazi2018mutual}, the authors use the Donsker-Varadhan~(DV)~\cite{donsker1983asymptotic} representation of the KL-divergence as the lower bound on the mutual information. Recently, another bound on mutual information, formulated as infoNCE~\cite{oord2018representation} based on Noise-Contrastive Estimation~\cite{gutmann2010noise}, has seen wide adoption in representation learning due to its low variance and accurate estimate of MI. It is defined as follows;

\begin{multline}
    \hat{I}_{\omega}^{InfoNCE}(X;Z) \\ := \mathbb{E}_{P(X, Z)}[T_{\omega} - \mathbb{E}_{P(X)P(Z)} [\log \sum T_{\omega}]],
\end{multline}


where $T_{\omega}: \mathcal{X} \times \mathcal{Z} \to \mathbb{R}$ is the discriminator neural network with parameters $\omega$. However, as demonstrated in~\cite{hjelm2018learning}, both DV and infoNCE require a large number of negative samples during training. Recent works tackle this problem by using a memory-buffer that keeps representations from previous samples in memory to be accessed during training. As implemented in CRD~\cite{tian2019contrastive}, this can be done if mutual information is maximized only between the final representations of the networks as the dimensions of the representations to be kept in memory is limited. In this work, we extend this infoNCE based MI maximization framework to include feature and local level information maximization. As a result, we require negative samples for each location in the multiple $K$ intermediate feature maps as well as for the final representations. This becomes unfeasible for most large state-of-the-art architectures. To this end, in our approach we adopt Jensen-Shannon divergence based mutual information estimation, similar to the formulations in~\cite{nowozin2016f} and~\cite{brakel2017learning}. The MI estimate from this JSD-based bound on MI, due to its formulation, is insensitive to the number of negative samples.

\begin{multline}~\label{eq:info}
    I(X;Z) \geq \hat{I}_{\omega}^{JSD}(X;Z) := \mathbb{E}_{P(X, Z)}[-\mathrm{log}(1 + e^{-T_{\omega}})] \\ 
    - \mathbb{E}_{P(X)P(Z)}[\mathrm{log}(1 + e^{T_{\omega}})].
\end{multline}

Overall, we optimize the parameters $\theta$ of the student network $f_s$ and parameters $\omega$ of the critic network $T_{\omega}$ by simultaneously estimating and maximizing mutual information between the representations of the frozen teacher network and the student network.

\subsection{Global information maximization} \label{sec:global}
Our global objective aims to maximize the mutual information between the richer final representation of the frozen teacher network $f_t(x)$ and the final representation of the student network $f_s(x)$ to encourage the student to learn richer representations. This objective uses a discriminator function $T_{\omega_g}$, where $\omega_g$ are the trainable parameters. We use the infoNCE bound for global MI maximization as it is computationally feasible to maintain a memory bank of negative samples due to the lower dimensionality of the final representations from the networks. We optimize the parameters of the student and the discriminator function simultaneously as: 

\begin{equation}
    (\hat{\omega_g}, \hat{\theta}) = \underset{\omega_g, \theta }{\mathrm{argmax}}\; \hat{I}^{\text{infoNCE}}_{\omega_g} (f_t(x), f_s(x)).
\end{equation}

\subsection{Local information maximization} \label{sec:local}
In this objective we maximize the mutual information between a richer final representation of the teacher network and representations of local regions extracted by the student network. This objective draws from the assertion that the final teacher representations contains valuable information required for downstream classification. Hence, this objective encourages the student network to extract information from local image regions that is ultimately useful for classification. 

We enforce this objective between $f_t(x)$ and the last intermediate representation from the student network in the set $\mathcal{R}$. Therefore for $k=K$, $f^{(K)}_s(x)$ is a $m_K \times m_K$ feature map where each location roughly corresponds to an $H/m_K \times W/m_K$ patch in the input image where $H, W$ are the height and width of the image. The representation of each such patch $\{f^{(K)}_s(x)\}_{i,j}$ is then paired with $f_t(x)$, where $i,j \in [1, m_K]$ denotes the specific location in the feature map. The pairs are then used with the mutual information estimator to optimize the parameters as follows:

\begin{equation}
    (\hat{\omega_l}, \hat{\theta}) = \underset{\omega_l, \theta }{\mathrm{argmax}}\; \frac{1}{m^2_K}  \sum_{i=1}^{m_K} \sum_{j=1}^{m_K} \hat{I}^{\text{JSD}}_{\omega_l} (f_t(x), \{f^{(K)}_s(x)\}_{i,j})
\end{equation}
where a discriminator neural network $T_{\omega_l}$ with parameters $\omega_l$ is used.

\subsection{Feature Information maximization} \label{sec:feature}
This objective aims to maximize the mutual information between region-consistent intermediate representations from the networks. In neural networks, the complexity of captured visual patterns increases towards the later layers~\cite{zeiler2014visualizing}. Intuitively, to mimic the representational power of the teacher, the student network needs to learn these complex patterns hierarchically. In order to motivate such hierarchical learning, mutual information is maximized between intermediate features at different depths in the networks. This enables the student to learn to identify complex patterns in a bottom-up fashion and systematically learn to generate richer features. Note that within each pair of intermediate feature maps in set $\mathcal{R}$, mutual information is maximized between vectors corresponding to the same location in the image. This information maximization pushes the student network to extract features from each region of the image that share maximum information with the features extracted by the teacher network from the same region. For a pair $(f_t^{(k)}(x), f_s^{(k)}(x)) \in \mathcal{R}$, information is maximized between pairs of region-consistent vectors $\{f^{(k)}_t(x)\}_{i,j}$ and $\{f^{(k)}_s(x)\}_{i,j}$ for each $i,j \in [1, m_k]$ as follows:

\begin{multline}
    (\hat{\omega_f}, \hat{\theta}) = \underset{\omega_f, \theta }{\mathrm{argmax}}\, \frac{1}{K}\frac{1}{m^2_k}  \sum_{k=1}^{K} \sum_{i=1}^{m_k} \sum_{j=1}^{m_k} \\  
    \hat{I}^{\text{JSD}}_{\omega_f} (\{f^{(k)}_t(x)\}_{i,j}, \{f^{(k)}_s(x)\}_{i,j})
\end{multline}

where a discriminator neural network $T_{\omega_f}$ with parameters $\omega_f$ is used.



\begin{figure*}[t]
\begin{center}
\includegraphics[width=0.80\linewidth]{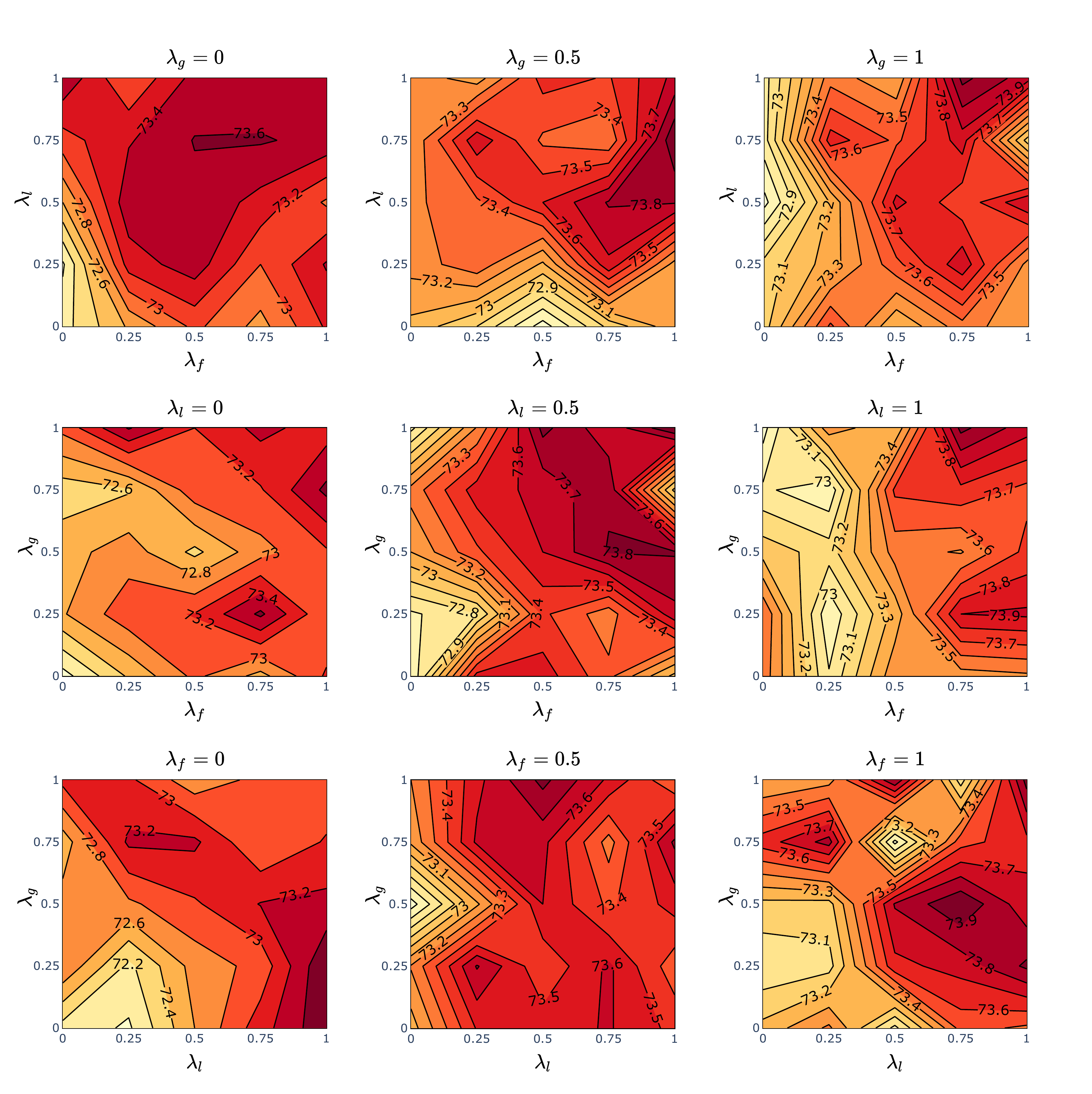}
\end{center}
   \caption{Results from the ablation studies on CIFAR100 dataset using a student ResNet-8x4 (baseline acc. $72.44\%$) with teacher ResNet-32x4 (baseline acc. $79.24\%$). Contour lines represent the final test accuracy of the student. The study was performed by varying the values of $\lambda_f$, $\lambda_g$, $\lambda_l$ from $0$ to $1$ with increments of $0.25$ while $\alpha$ was kept constant at $1$. In each plot, the accuracy landscape is shown with $\lambda_g$ set to a constant value. Plots for remaining values of $\lambda$ have been added to the appendix. }
\label{fig:contour}
\end{figure*}

\subsection{Classification objective}
Here the cross-entropy loss is minimized between the output of the classification function $f_s^{cls}(x)$ and the target label $y$ as follows:

\begin{equation}
    (\hat{\theta}) = \underset{\theta} {\mathrm{argmin}}\; \mathcal{L}_{CE} (y, f_s^{cls}(x)),
\end{equation}
where $\mathcal{L}_{CE}$ denotes the cross-entropy function. 

\noindent
Our overall objective is a weighted-summation of all the above individual objectives with weights $\alpha$ (cross-entropy loss), $\lambda_g$ (global MI maximization), $\lambda_l$ (local MI maximization), and $\lambda_f$ (feature MI maximization)

\subsection{Mutual Information Discriminators}
The parameterized mutual information discriminator functions ($T_{\omega_g}$, $T_{\omega_l}$, and $T_{\omega_f}$) can be modeled as neural networks. In our experiments, we use two distinct discriminator architectures inspired from the functions presented in Deep InfoMax~\cite{hjelm2018learning}. For global information maximization, we use the standard project and dot architecture. The representations from both the teacher and the student are first projected using an appropriate projection architecture with a linear shortcut. The dot-product of these projections is then computed to get the score. Positive and negative pairs of representations are passed through the discriminator to get respective scores to be passed into equation~\ref{eq:info} to get the estimates on the lower bound of the mutual information. Whereas, for local and feature information maximization we use a convolution based architecture as it is cheaper for higher dimensional inputs. \\

\begin{wraptable}{}{0.5\columnwidth}
\vspace{-0.1in}
\centering	
 \resizebox{0.5\columnwidth}{!}{%
	\begin{tabular}{|c|c|c|} \hline
    Input & Operation  & Output \\ \hline
    $[f_t(x), f^{(K)}_s(x)]$ & $1 \times 1$ Conv + ReLU & $O_1$ \\ 
    $O_1$ & $1 \times 1$ Conv + ReLU & $O_2$ \\
    $O_2$ & $1 \times 1$ Conv & scores \\
    \hline
    \end{tabular}}
\end{wraptable}

\noindent
Specifically, for local information maximization, we replicate the final representation from the teacher $f_t(x)$ to match the $m_K \times m_K$ size of the student's last intermediate feature map ($f^{(K)}_s(x)$). The resulting replicated tensor is then concatenated with $f^{(K)}_s(x)$ to get $[f_t(x), f^{(K)}_s(x)]$ which serves as the input for the critic function (ref. table on right). Similarly, consider feature mutual information maximization, for each pair in the set $\mathcal{R}$ we use a distinct discriminator $T^{(k)}_{\omega_f}$. For a given $k$, each pair of intermediate feature representations in the set $\mathcal{R}$ are concatenated together to get $[f^{(k)}_t(x), f^{(k)}_s(x)]$. Which is then passed through two convolutional ($1\times1$ kernels and $512$ filters) where each layer is followed by a ReLU non-linearity. The output obtained is then further passed into a convolutional layer ($1\times1$ kernels and $1$ filter) to give $m_k \times m_k$ scores (ref. table on right). Further details are provided in supplementary.

\begin{table*}[t]

\newcolumntype{Y}{>{\raggedright\arraybackslash}X}
\newcolumntype{Z}{>{\centering\arraybackslash}X}
\newcolumntype{W}[1]{>{\centering\arraybackslash\hspace{0pt}}p{#1}}

\centering
\setlength\tabcolsep{0.1pt}
\renewcommand{\arraystretch}{1.5}
\begin{tabularx}{\textwidth}{l p{0.15in} YYYYYYY}
\toprule
{\bf Student Net.}          &&
{\bf WRN-16-1} & {\bf WRN-16-2} & {\bf ResNet-8}   & {\bf ResNet-20}  & {\bf ResNet-20} & {\bf ResNet-8x4}  & {\bf VGG-8}  \\ 

{\bf Teacher Net.}          &~~~& 
{\bf WRN-40-2} & {\bf WRN-40-2} & {\bf ResNet-110} & {\bf ResNet-110} & {\bf ResNet-56} & {\bf ResNet-32x4} & {\bf VGG-19} \\
\midrule

\band
Student Acc. && 
67.01    & 72.80    &  59.63    &    69.10  &  69.10   & 72.44      & 69.67 \\ 

Teacher Acc. && 
75.31\RiseB{8.30}    & 75.31\RiseB{2.51}    &  73.82\RiseB{14.19}    &    73.82\RiseB{4.72}  &  72.31\RiseB{3.21}   & 79.24\RiseB{3.80}      & 74.63\RiseB{4.96} \\

\midrule

FitNets &&
68.35\Rise{1.34}    & 73.11\rise{0.31}    & 60.36\rise{0.73}     & 69.12\rise{0.02}     & 69.28\rise{0.18}    & 73.80\Rise{1.36}      & 71.32\Rise{1.65} \\

AT &&
68.49\Rise{1.48}    & 73.37\rise{0.57}    & 60.24\rise{0.61}     & 70.36\Rise{1.26}     & 70.18\Rise{1.08}    & 73.20\rise{0.76}      & 71.71\Rise{2.04} \\ 
\midrule

VID &&
68.95\Rise{1.94}    & 73.89\Rise{1.09}    & 60.44\rise{0.81}     & 70.32\Rise{1.22}     & 70.52\Rise{1.42}    & 73.19\rise{0.75}      & 71.52\Rise{1.85} \\

KD && 
68.24\Rise{1.23}    & 73.91\Rise{1.11}    & 61.01\Rise{1.38}     & 70.32\Rise{1.22}     & 70.59\Rise{1.40}    & 73.21\rise{0.77}      & 72.29\Rise{2.62} \\

CRD &&
69.21\Rise{2.20}    & 74.17\Rise{1.37}    & 60.82\Rise{1.19}     & \textbf{71.45}\Rise{2.35} & 71.12\Rise{2.02}    & 75.21\Rise{2.77}      & 73.10\Rise{3.43} \\ 
\midrule

MIMKD (ours)   && 
\textbf{70.20}\Rise{3.19} & \textbf{75.16}\Rise{2.36} & \textbf{61.81}\Rise{2.18}  & 71.43\Rise{2.33}     & \textbf{71.31}\Rise{2.21} & \textbf{75.83}\Rise{3.39} & \textbf{73.27}\Rise{3.60}  \\ 
\bottomrule

\end{tabularx}

\caption{\label{tab:same} Observed test accuracy (in \%) of student networks trained with teacher networks of higher capacity but similar architecture on the CIFAR100 dataset using MIMKD and other competing methods. MIMKD shows consistent increases in accuracy for all model pairs and the largest gains overall.}
\end{table*}

\subsection{Implementation Details}
We adopted the generally established approach for training CNNs on the CIFAR-100 dataset. We use SGD with momentum $0.9$, weight decay $5 \times 10^{-4}$, and an initial learning rate of $0.05$ for a total of $240$ epochs with batch-size $64$. The learning rate is decayed by $0.1$ at the $150$th, $180$th and the $210$th epoch. We used random horizontal flips and random crop for augmenting the dataset during training. For ImageNet, we use the standard PyTorch training scheme for ResNets~\cite{he2016deep}. Code implementation will be made public on publication.

%% file: src/040experiments.tex
In this section, we demonstrate the efficacy of our framework using various ablative and quantitative analyses. We first establish the value of each of our mutual information maximization formulations by performing an extensive ablative study (sec.~\ref{sec:ablation}). Further, we demonstrate the prowess of our distillation framework based on model compression performance in the following setups: (1) Under similar student-teacher network architectures (sec.~\ref{sec:similar}), (2) under dissimilar architectures (sec.~\ref{sec:dissimilar}), (3) under a setting with custom designed shallow student networks (ref. appendix for results), (4) in a larger scale setting on Imagenet (ref. appendix for results), and (5) in terms of transfer learning performance (sec.~\ref{sec:transfer}) as a measure of the transferability of distilled representations. Our model compression experiments are performed on the CIFAR-100 dataset which contains colored natural images of size $32 \times 32$. It has $50$K training images with $500$ images in each of $100$ classes and a total of $10$K test images. In our experiments, we use standard CNN architectures of varied capacities, such as ResNet~\cite{he2016deep}, Wide ResNet (WRN)~\cite{zagoruyko2016wide}, MobileNet~\cite{sandler2018mobilenetv2}, ShuffleNet~\cite{zhang2018shufflenet}, and VGG~\cite{simonyan2014very}. We compare our method with other knowledge distillation methods, such as (1) Knowledge Distillation (KD)~\cite{hinton2015distilling}, (2) FitNets~\cite{romero2014fitnets}, (3) Attention Transfer (AT)~\cite{zagoruyko2016paying}, (4) Variational Information Distillation (VID)~\cite{ahn2019variational}, and (5) Contrastive Representation Distillation (CRD)~\cite{tian2019contrastive}. We used the following values for hyper-parameters based on a held out set: $\alpha = 1$, $\lambda_g=1$, $\lambda_l=0.75$, $\lambda_f=1$ for all our experiments. The infoNCE bound in CRD as well as our global MI is set to use $4096$ negatives. The hyper-parameter choice for other approaches can be found in supplementary. Additionally, in order to demonstrate the scalability of our method, we compare our distillation performance on the ImageNet~\cite{deng2009imagenet} dataset against AT~\cite{zagoruyko2016paying}, and KD~\cite{hinton2015distilling}. ImageNet is a large-scale dataset with $1.2$ million training images across $1$K classes and a total of $50$K validation images. 


\subsection{Ablation Study}
\label{sec:ablation}
We perform an extensive ablation study to demonstrate the value of each component of our mutual information maximization objective. Ablative study experiments are performed with ResNet-32x4 as the teacher network and ResNet-8x4 as the student network where the baseline accuracy of the teacher is $79.24 \%$ and that of the student network is $72.44 \%$. The values of the hyper-parameters $\lambda_g$, $\lambda_l$ and $\lambda_f$ --- that control the weight of the global, local and feature mutual information maximization objectives respectively -- were varied between $0$ and $1$ with an increment of $0.25$ while the weight for the cross-entropy loss, $\alpha$ was set to $1$. Note that for this study, we use the JSD-based bound for all MI maximization formulations including for global MI which is not the case for our final competitive models presented further. The contour plots in Figure \ref{fig:contour} shows the test accuracy landscape with respect to a pair of hyper-parameters when the third hyper-parameter is set to distinct values. For instance, we observe that for any value of $\lambda_g$, better performance is achieved towards higher values of both $\lambda_f$ and $\lambda_l$. Similar trends can be observed in all the accuracy landscape plots. Overall, this demonstrates the value of maximizing region-consistent local and feature-level mutual information between representations in addition to just global information maximization. Please refer to the appendix for additional accuracy landscape plots.

\begin{table*}[t]

\newcolumntype{Y}{>{\raggedright\arraybackslash}X}
\newcolumntype{Z}{>{\centering\arraybackslash}X}
\newcolumntype{W}[1]{>{\centering\arraybackslash\hspace{0pt}}p{#1}}

\centering
\setlength\tabcolsep{0.1pt}
\renewcommand{\arraystretch}{1.5}

\begin{tabularx}{\textwidth}{l p{0.1in} YYYYYY}
\toprule

{\bf Student Net.}          &&
{\bf WRN-16-1}  & {\bf WRN-16-2}   & {\bf VGG-8}       & {\bf ShuffleNetV1} & {\bf ShuffleNetV2} & {\bf MobileNetV2} \\ 

{\bf Teacher Net.}          &~~~~~& 
{\bf ResNet-110} & {\bf ResNet-32x4} & {\bf ResNet-32x4} & {\bf VGG-13}        & {\bf ResNet-50}     & {\bf VGG-13}       \\
\midrule
\band
Student Acc. &&
67.01     & 72.80      & 69.67      & 70.51        & 69.85        & 61.11       \\ 

Teacher Acc. &&
73.82\RiseB{6.81}     & 79.24\RiseB{6.44}      & 79.24\RiseB{9.57}      & 74.62\RiseB{4.11}        & 79.23\RiseB{9.38}        & 74.62\RiseB{13.51}       \\
\midrule

FitNets &&
67.99\rise{0.98}     & 73.79\rise{0.99}      & 70.28\rise{0.61}      & 72.29\Rise{1.78}        & 71.80\Rise{1.95}        & 61.42\rise{0.31}       \\

AT &&
66.42\drop{0.59}     & 72.19\drop{0.61}      & 71.77\Rise{2.10}      & 71.19\rise{0.68}        & 70.78\rise{0.93}        & 61.96\rise{0.85}       \\ 
\midrule

VID &&
67.47\rise{0.46}     & 73.38\rise{0.58}      & 71.52\Rise{1.85}      & 72.22\Rise{1.71}        & 72.84\Rise{2.99}        & 63.01\Rise{1.90}       \\

KD &&
68.86\Rise{1.85}     & 74.63\Rise{1.83}      & 73.46\Rise{3.79}      & 72.26\Rise{1.75}        & 72.91\Rise{3.06}        & 64.47\Rise{3.36}       \\

CRD &&
69.71\Rise{2.70}     & 75.61\Rise{2.81}      & 73.73\Rise{4.06}      & 72.86\Rise{2.35}        & 73.65\Rise{3.80}        & \textbf{66.34}\Rise{5.23} \\ 
\midrule

MIMKD (ours)  &&
\textbf{69.88}\Rise{2.87} & \textbf{76.24}\Rise{3.44} & \textbf{74.09}\Rise{4.42} & \textbf{73.88}\Rise{3.37} & \textbf{74.55}\Rise{4.70} & 65.89\Rise{4.78}  \\

\bottomrule
\end{tabularx}

\caption{\label{tab:diff} Observed test accuracy (in \%) of student networks trained with teacher networks of higher capacity and different architecture on the CIFAR100 dataset using our method MIMKD and other distillation frameworks.}
\vspace{-0.15in}
\end{table*}

\subsection{Similar CNN Architectures}
\label{sec:similar}
We perform knowledge distillation from a teacher network to a student network of the same family (e.g. ResNets of different capacities). Table \ref{tab:same} presents our results, showing that our method outperforms others in most setups and always obtains gains with respect to student networks. Notice that CRD~\cite{tian2019contrastive} is able to slightly surpass the performance of our method in one setup while being close in most cases. We find this encouraging as CRD~\cite{tian2019contrastive} uses a similar mutual information maximization based formulation in their distillation objective with a tighter lower-bound. Therefore, if we only use the global objective in our method, CRD~\cite{tian2019contrastive} should outperform our method due to its tighter bound. Despite compromising the lower bound on mutual information, MIMKD takes advantage of using region-consistent local and feature-level mutual information maximization.

\subsection{Dissimilar CNN Architectures}
\label{sec:dissimilar}
Here, we perform knowledge distillation from a teacher network to a student network with a significantly different architecture. This tests the flexibility methods to adapt to distinct data-abstraction flows of dissimilar neural network architectures. Table \ref{tab:diff} demonstrates that our method (MIMKD) outperforms other distillation methods in most teacher-student combinations increasing the accuracy of a ShuffleNetV2 by 4.7\% while distilling from a much different ResNet-50 model. This demonstrates that our method is able to accommodate significant architectural differences in teacher-student pairs and does not impose structural constraints on intermediate layers that hinder training. While other methods that work on intermediate feature maps like AT~\cite{zagoruyko2016paying} and FitNets~\cite{romero2014fitnets} do not show much improvement from base student accuracy.

\subsection{Transferring representations}
\label{sec:transfer}
Finally, we compare the transferability of the features learned with knowledge distillation from MIMKD and baselines, on two other datasets: STL-10 and TinyImagenet. A WRN-16-2 network is trained with and without distillation from a pre-trained WRN-40-2 teacher on the CIFAR100 dataset. The student is then used as a frozen feature extractor (pre-classification layer) for images in the STL-10 and the TinyImageNet dataset. A linear classifier is trained on these extracted features to perform classification on the test sets of these datasets. The classification accuracy on the unseen datasets is interpreted as the transferability of representations. Results are presented in Table~\ref{tab:transfer} and show that MIMKD learns more transferrable representations.

\begin{table}[h]
\small

\newcolumntype{Y}{>{\raggedright\arraybackslash}X}
\newcolumntype{Z}{>{\centering\arraybackslash}X}
\setlength\tabcolsep{1pt}
\renewcommand{\arraystretch}{1.0}
\centering
\begin{tabularx}{\columnwidth}{l c ZZ}
\toprule
         &~& {\bf STL-10} & {\bf TinyImageNet} \\ 
\midrule
\band
Base Accuracy { (no distillation)} && 69.5   & 33.8         \\ 
\midrule

{ Knowledge~Distillation~}(KD)       && 70.6   & 33.9         \\
{ Attention~Transfer~}(AT)       && 70.8   & 34.4         \\
{ Contrastive~Repr.~Distill~}(CRD)      && 71.4   & 35.6         \\ 
\midrule

MIMKD (this work)    && \textbf{71.8}   & \textbf{36.2}\\
\bottomrule

\end{tabularx}
\caption{\label{tab:transfer} Observed test-set accuracy (in \%) of the student network on STL-10 and TinyImagenet datasets using our method (MIMKD) and other distillation frameworks.}
\vspace{-0.2in}
\end{table}

%% file: src/050discussion.tex
In this paper, we presented a framework (MIMKD) motivated by an information-theoretic perspective on knowledge distillation. Utilizing an information-efficient lower bound on mutual information, we proposed three information maximization formulations and demonstrated the value of region-consistent local and feature-level information maximization on distillation. We enable intermediate distillation using a JSD based lower-bound on MI which we optimize using only one negative sample. Further works in this area could explore our contention that if used with a tighter lower-bound, our feature and local information maximization objectives have the potential to surpass even its current performance. 

\section{Acknowledgements}
This work was supported by NSF Awards IIS-2221943 and IIS-2201710, and through gift funding from a Facebook Research Award: Towards On-Device AI.

%% file: src/090appendix.tex
\section{Appendix}

\subsection{Limitations and Broader Impacts}
\begin{figure*}
\begin{center}
\includegraphics[width=0.95\linewidth]{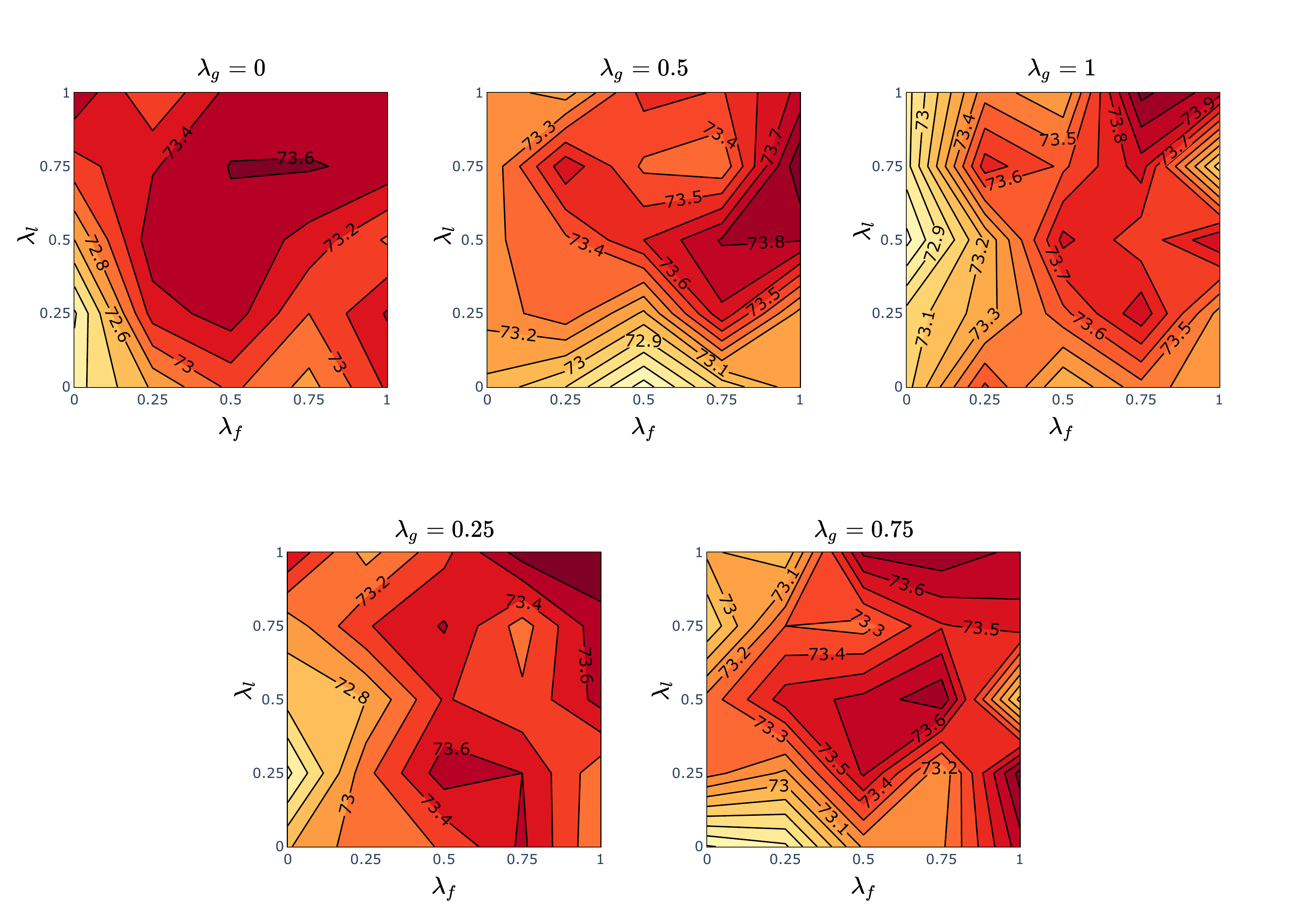}
\end{center}
   \caption{Results from the ablation studies on CIFAR100 dataset using a student resnet8x4 (baseline acc. $72.44\%$) with teacher resnet32x4 (baseline acc. $79.24\%$). Contour lines represent the final test accuracy of the student. Grid search was performed by varying the values of $\lambda_f$, $\lambda_g$, $\lambda_l$ from $0$ to $1$ with increments of $0.25$. In each plot, the accuracy landscape is shown with $\lambda_g$ set to a constant value.}
\label{fig:contour1}
\end{figure*}

\begin{figure*}
\begin{center}
\includegraphics[width=0.95\linewidth]{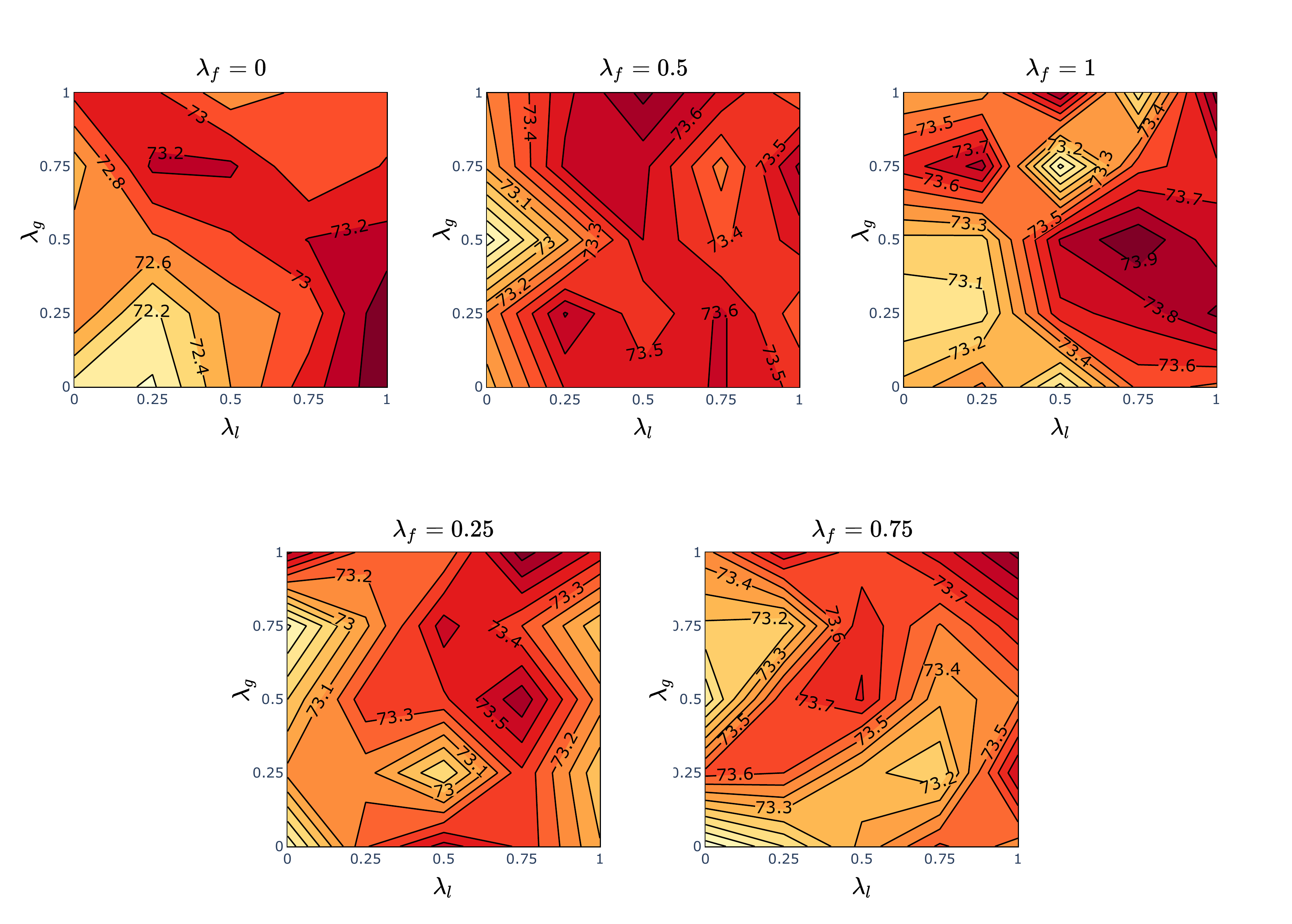}
\end{center}
   \caption{Results from the ablation studies on CIFAR100 dataset using a student resnet8x4 (baseline acc. $72.44\%$) with teacher resnet32x4 (baseline acc. $79.24\%$). Contour lines represent the final test accuracy of the student. Grid search was performed by varying the values of $\lambda_f$, $\lambda_g$, $\lambda_l$ from $0$ to $1$ with increments of $0.25$. In each plot, the accuracy landscape is shown with $\lambda_f$ set to a constant value.}
\label{fig:contour2}
\end{figure*}

\begin{figure*}
\begin{center}
\includegraphics[width=0.95\linewidth]{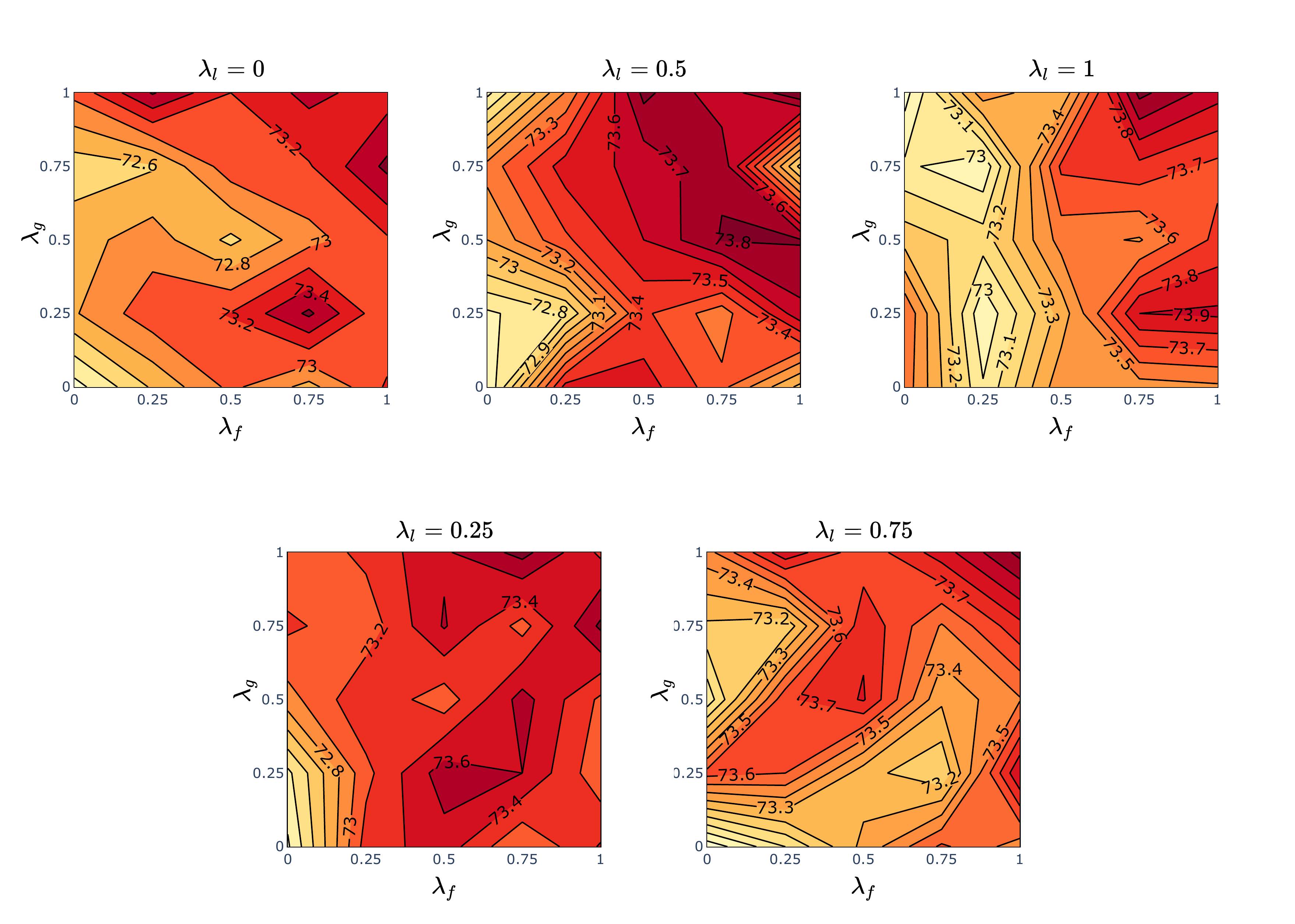}
\end{center}
   \caption{Results from the ablation studies on CIFAR100 dataset using a student resnet8x4 (baseline acc. $72.44\%$) with teacher resnet32x4 (baseline acc. $79.24\%$). Contour lines represent the final test accuracy of the student. Grid search was performed by varying the values of $\lambda_f$, $\lambda_g$, $\lambda_l$ from $0$ to $1$ with increments of $0.25$. In each plot, the accuracy landscape is shown with $\lambda_l$ set to a constant value.}
\label{fig:contour3}
\end{figure*}

In this paper, we presented a novel Mutual Information Maximization based knowledge distillation framework (MIMKD). Our method uses the JSD based lower-bound on mutual information which is optimized using only one negative sample. However, despite its favorable properties, our lower-bound may be less tight on the mutual information than the infoNCE bound as it approximates the mutual information by being monotonically related with it. Additionally, as we use only one negative sample, the performance of the method may be hindered by the presence of false negatives. The performance of the method is also effected by the architecture of the discriminator functions which can be explored further. We presented three information maximization formulations and demonstrated the value of region-consistent information maximization on distillation performance. We observe that the performance is slightly-sensitive to the hyper-parameters that control the relative value of our global, local, and feature information maximization formulations. This has been explored in great detail in our ablation sections and further demonstrated in figures~\ref{fig:contour1},~\ref{fig:contour2}, and ~\ref{fig:contour3}. Our method transfers representations from the teacher to the student. As such, harmful biases that the teacher has learnt are transferred to the student as well. And further exploration is required to alleviate the transfer of such biases during distillation.

\subsection{Hyper-parameters for other methods}

The student is trained with the following loss function which is a combination of the distillation loss and the cross-entropy loss for classification:

\begin{equation}
    \mathcal{L} = \alpha \mathcal{L}_{cls} + (1 - \alpha) \mathcal{L}_{KD} + \beta \mathcal{L}_{dis} 
\end{equation}

Note that we set $\alpha=1$ for all methods except KD~ß\cite{hinton2015distilling} and the value of $\beta$ is set to the value recommended in the original work as follows:

\begin{enumerate}
    \item KD \cite{hinton2015distilling}: $\alpha = 0.9, \beta = 0$
    \item Fitnet \cite{romero2014fitnets}: $\beta = 100$
    \item AT \cite{zagoruyko2016paying}: $\beta = 1000$
    \item VID \cite{ahn2019variational}: $\beta = 1$
    \item CRD \cite{tian2019contrastive}: $\beta = 0.8$, for CRD evaluation, we use a original work inspired self-implementation with $4096$ negative samples and $i \neq j$ negative sampling methodology as described in the original work. 
\end{enumerate}

\subsection{Pairing Intermediate Representations}
\subsubsection{Similar CNN Architectures.}
 
 Consider the case of distillation when the teacher network is a pre-trained WRN-40-2 and the student network is a WRN-16-1. We use $4$ same-sized representations extracted from intermediate layers of the networks. Therefore, the set  $\mathcal{R} = \{(f_t^{(k)}(x), f_s^{(k)}(x))\}_{k=1}^{K}$ contains $k$ pairs of same-sized 2-dimensional representations. Table \ref{tab:same_maps} describes the sizes of the intermediate representations used for feature-based mutual information maximization. It can be seen that for this combination we use $k=4$ in our formulation. 

\begin{table}[htb]
\caption{\label{tab:same_maps} Dimensions of intermediate representation in the form $channels \times height \times width$ used for feature-level mutual information maximization between a teacher WRN-40-2 and a student WRN-16-1 network. Alternatively, each value of $k$ represents a pair of elements in the set $\mathcal{R}$.}
\begin{center}
\begin{tabular}{|l|c|c|} \hline
  & WRN-40-2               & WRN-16-1               \\ \hline
k & $f_t^{(k)}(x)$         & $f_s^{(k)}(x)$         \\ \hline
1 & 16 $\times$ 32 $\times$ 32 & 16 $\times$ 32 $\times$ 32 \\
2 & 32 $\times$ 32 $\times$ 32 & 16 $\times$ 32 $\times$ 32 \\
3 & 64 $\times$ 16 $\times$ 16 & 32 $\times$ 16 $\times$ 16 \\
4 & 128 $\times$ 8 $\times$ 8  & 64 $\times$ 8 $\times$ 8 \\
\hline
\end{tabular}
\end{center}
\end{table}

\subsubsection{Dissimilar CNN Architectures.}

Similar approach of defining the set $\mathcal{R}$ is followed in cases where the teacher and student networks have significantly different architectures. For instance, Table \ref{tab:diff_maps} shows the dimensions of intermediate representations used when the teacher network is a ResNet34 while the student is a ShuffleNetV2. Here $k=4$ is used, however, for some combinations of different standard architectures we use $k=3$ if only $3$ pairs intermediate representations from the teacher and the student have the same size. Note that our method is invariant to the number of channels in the representations. Therefore, mismatch in the number of channels in pairs of representations in $\mathcal{R}$ is inconsequential for the formulation of our losses. 

\begin{table}[htb]
\begin{center}
\caption{\label{tab:diff_maps} Dimensions of intermediate representation in the form $channels \times height \times width$ used for feature-level mutual information maximization between a teacher WRN-40-2 and a student WRN-16-1 network. Alternatively, each value of $k$ represents a pair of elements in the set $\mathcal{R}$.}
\begin{tabular}{|l|c|c|} \hline
  & ResNet34               & ShuffleNetV2               \\ \hline
k & $f_t^{(k)}(x)$         & $f_s^{(k)}(x)$         \\ \hline
1 & 64 $\times$ 32 $\times$ 32 & 24 $\times$ 32 $\times$ 32 \\
2 & 512 $\times$ 16 $\times$ 16 & 116 $\times$ 16 $\times$ 16 \\
3 & 1024 $\times$ 8 $\times$ 8 & 232 $\times$ 8 $\times$ 8 \\
4 & 2048 $\times$ 4 $\times$ 4 & 464 $\times$ 4 $\times$ 4 \\
\hline
\end{tabular}
\end{center}
\end{table}

\subsection{Mutual Information Discriminators}
The parameterized mutual information discriminator functions ($T_{\omega_g}$, $T_{\omega_l}$, and $T_{\omega_f}$) can be modeled as neural networks. In our experiments, we use two distinct discriminator architectures inspired from the functions presented in Deep InfoMax~\cite{hjelm2018learning}.

\subsubsection{Convolve Architecture.} In this method, the representations from the teacher and the student are concatenated together and passed through a series of layers to get the score. For global information maximization, the final representations from both networks is concatenated together to get $[f_s(x), f_t(x)]$. This vector is then passed to a fully connected network with two $512$-unit hidden layers, each followed by a $ReLU$ non-linearity (ref. table \ref{tab:g_disc}). The output is then passed through another linear layer to obtain the final score. 

\begin{table}[htb]
\begin{center}
\caption{\label{tab:g_disc} The architecture of the discriminator used for global information maximization. Here LL denotes Linear Layer and $d(v)$ refers to the number of dimensions in vector $v$.}
\begin{tabular}{|c|c|c|} \hline
Input & Operation  & Output \\ \hline
$[f_t(x), f_s(x)]$ & LL + ReLU & $O_1$ \\ 
$O_1$ & LL + ReLU & $O_2$ \\
$O_2$ & LL & score \\
\hline
\end{tabular}
\end{center}
\end{table}

For local information maximization, we replicate the final representation from the teacher $f_t(x)$ to match the $m_K \times m_K$ size of the student's last intermediate feature map ($f^{(K)}_s(x)$). The resulting replicated tensor is then concatenated with $f^{(K)}_s(x)$ to get $[f_t(x), f^{(K)}_s(x)]$ which serves as the input for the critic function (ref. table \ref{tab:lf_disc}).

\begin{table}[htb]
\begin{center}
\caption{\label{tab:lf_disc} The architecture of the discriminator used for local and feature mutual information maximization. Note that for feature mutual information maximization the input at the first layer is $[f^{(k)}_t(x), f^{(k)}_s(x)]$.}
\begin{tabular}{|c|c|c|} \hline
Input & Operation  & Output \\ \hline
$[f_t(x), f^{(K)}_s(x)]$ & $1 \times 1$ Conv + ReLU & $O_1$ \\ 
$O_1$ & $1 \times 1$ Conv + ReLU & $O_2$ \\
$O_2$ & $1 \times 1$ Conv & scores \\
\hline
\end{tabular}
\end{center}
\end{table}

Similarly, consider feature mutual information maximization, for each pair in the set $\mathcal{R}$ we use a distinct discriminator $T^{(k)}_{\omega_f}$. For a given $k$, each pair of intermediate feature representations in the set $\mathcal{R}$ are concatenated together to get $[f^{(k)}_t(x), f^{(k)}_s(x)]$. Which is then passed through two convolutional ($1\times1$ kernels and $512$ filters) where each layer is followed by a ReLU non-linearity. The output obtained is then further passed into a convolutional layer ($1\times1$ kernels and $1$ filter) to give $m_k \times m_k$ scores (ref. table \ref{tab:lf_disc}).

\subsubsection{Project and Dot Architecture.}
In this method, the representations from both the teacher and the student are first projected using an appropriate projection architecture with a linear shortcut. The dot-product of these projections is then computed to get the score. Positive and negative pairs of representations are passed through the discriminator to get respective scores to be passed into equation (2) to get the estimates on the lower bound of the mutual information. One-dimensional representations are projected using the architecture described in table \ref{tab:proj_disc}, whereas for two-dimensional intermediate feature maps, projection architecture described in table \ref{tab:proj_disc_2} is used.

\begin{table}[htb]
\begin{center}
\caption{\label{tab:proj_disc} The projection architecture used for one-dimensional inputs.  Here, LL denotes linear layer while LN denotes layer normalization. Both $f_t(x)$ and $f_s(x)$ are projected using this architecture and their dot product is computed to get scores.}
\begin{tabular}{|c|c|c|} \hline
Input & Operation  & Output \\ \hline
$f_t(x)$ or $f_s(x)$ & LL + ReLU + LL & $O_1$ \\
$f_t(x)$ or $f_s(x)$ & LL + ReLU & $O_2$ \\
$O_1$ + $O_2$ & LN & $proj$ \\
\hline
\end{tabular}
\end{center}
\end{table}

Therefore, for (1) global information maximization, both $f_t(x)$ and $f_s(x)$ are projected using the one-dimensional projection architecture, for (2) local information maximization, the final teacher representation, $f_t(x)$, is projected using the one-dimensional projection architecture and duplicated to match the size of the projected intermediate student representation projected using the two-dimensional projection architecture, a dot product of these outputs is then computed to get the scores, while for (3) feature information maximization, both representations in each pair of the set $\mathcal{R}$ is projected using a respective two-dimensional projection architecture. 

\begin{table}[htb]
\begin{center}
\caption{\label{tab:proj_disc_2} The projection architecture used for two-dimensional inputs.  Here, LL denotes linear layer while LN denotes layer normalization.}
\begin{tabular}{|c|c|c|} \hline
Input & Operation  & Output \\ \hline
$f^{(k)}_s(x)$ & $1 \times 1$ Conv + ReLU + LL & $O_1$ \\
$f^{(k)}_s(x)$ & $1 \times 1$ Conv + ReLU & $O_2$ \\
$O_1$ + $O_2$ & LN & $proj$ \\
\hline
\end{tabular}
\end{center}
\end{table}

\begin{table}[]
\caption{\label{tab:imagenet} Observed top-1 validation accuracy (in \%) of the student network on the ImageNet dataset using our method (MIMKD) and other distillation frameworks. In similar settings, the more recent Contrastive Representation Distillation (CRD) method reports comparable performance with an improvement of $+1.42$ from a student network~\cite{tian2019contrastive}.}
\centering
\begin{tabular}{lcc}
\toprule
{\bf Student Network}  && {\bf ResNet-18} \\ 
{\bf Teacher Network}  && {\bf ResNet-34} \\
\midrule

Student Accuracy             && 68.88    \\ 
Teacher Accuracy             && 72.82\RiseB{3.94}     \\
\midrule

Knowledge~Distill.~(KD) && 69.66\rise{0.78}   \\
Attention~Transfer~(AT) && 69.70\rise{0.82}   \\ 
\midrule

MIMKD (this work) && \textbf{70.32}\Rise{1.44}   \\
\bottomrule

\end{tabular}
\end{table}

\subsection{ImageNet results}
\label{sec:imagenet}
In this experiment we train a student ResNet-18 with a pre-trained teacher ResNet-34 on the ImageNet dataset (ILSVRC). Note that we do not perform any hyper-parameter tuning specifically for this configuration and use the same values we obtained for the CIFAR-100 dataset i.e. $\alpha = 0.9$, $\lambda_g=0.2$, $\lambda_l=0.8$, $\lambda_f=0.8$. We observed that our method is able to reduce the gap between the teacher and the student performance by~$1.44\%$. Results are presented in Table~\ref{tab:imagenet}. 

\subsection{Shallow CNN Architectures}
\label{sec:shallow}
In this section, we describe our experiments where we distill knowledge from a standard teacher network into a shallow custom-designed CNN. This is done to demonstrate that it is feasible to design and distill information into light-weight models such that they perform competitively with standard CNN architectures while running faster. For our experiments we use 2 shallow CNNs; (1) Conv-4 with 4 convolutional-blocks followed by average pooling operation and a linear layer, where each convolutional-block is made-up of a convolutional layer with kernel size $3 \times 3$ and stride $2$ followed by batch-normalization and a ReLU non-linearity, (2) Conv-4-MP which has 4 convolutions blocks followed by average pooling and a linear layer at the end, where each convolutional-block contains a convolutional layer with kernel size $3 \times 3$ and stride $1$ followed by batch-normalization, ReLU and a max-pooling layer. These architectures were chosen as they are compact and run relatively faster on standard CPUs. Table \ref{tab:shallow} compiles our results compared to other distillation methods for custom-designed shallow CNN architectures. Notice how a simple model such as Conv-4-MP becomes competitive with ShuffleNetV2's base student accuracy. Our method is able to outperform all other methods in this setup. Additionally, we can see that distillation is most successful with ResNet-32x4 as the teacher than for other architectures. This could be because of the larger gap in the baseline accuracy of the networks. Under this more controlled experiment with fixed students, larger gaps between student-teacher pairs also led to larger gains after distillation.

\begin{table*}[h]
\caption{\label{tab:shallow} Observed test accuracy (in \%) of shallow student networks trained with teacher networks of higher capacity and standard architectures on the CIFAR100 dataset using our methods MIMKD and other distillation frameworks.}

\newcolumntype{Y}{>{\raggedright\arraybackslash}X}

\centering
\setlength\tabcolsep{0.1pt}
\renewcommand{\arraystretch}{1.5}

\begin{tabularx}{\textwidth}{l p{0.1in} YYY p{0.02in} YYY}
\toprule

{\bf Student Net.} &~~~~~&
\multicolumn{3}{c}{\bf Conv-4} &~~~~~& \multicolumn{3}{c}{\bf Conv-4-MP}  \\

\cmidrule{3-5}\cmidrule{7-9}

{\bf Teacher Net.} &&
{\bf ResNet-110}  & {\bf VGG-13}  & {\bf ResNet-32x4} && 
{\bf ResNet-110}  & {\bf VGG-13} & {\bf ResNet-32x4}\\ 
\midrule

\band
Student Acc. &&
59.97     & 59.97      & 59.97  && 66.09     & 66.09      & 66.09 \\ 

Teacher Acc. &&
73.82\RiseB{13.85}      & 74.62\RiseB{14.65}  & 79.24\RiseB{19.27} &&
73.82\RiseB{7.73}       & 74.62\RiseB{8.53} & 79.24\RiseB{13.15}\\
\midrule

FitNets &&
60.58\rise{0.61}   & 61.81\Rise{1.84}    & 62.89\Rise{2.92}&&
67.38\Rise{1.29}   & 66.52\rise{0.43} & 67.21\Rise{1.12}\\

AT &&
61.65\Rise{1.68}     & 62.16\Rise{2.19}   & 63.10\Rise{3.13}&&
67.52\Rise{1.43}     & 66.21\rise{0.12} & 66.03\drop{0.06}\\ 
\midrule

VID &&
61.93\Rise{1.96}     & 62.49\Rise{2.52}   & 63.45\Rise{3.48}&&
67.76\Rise{1.67}     & 67.40\Rise{1.31} & 67.86\Rise{1.77}\\

KD &&
61.98\Rise{2.01}   & 62.10\Rise{2.13} & 62.87\Rise{2.90} &&
67.51\Rise{1.42}      & 67.84\Rise{1.75} & 68.04\Rise{1.95}\\

CRD &&
62.13\Rise{2.16}     & 62.54\Rise{2.57}  & 63.76\Rise{3.79}&&
67.96\Rise{1.87}    & 68.06\Rise{1.97} & 68.52\Rise{2.43} \\ 
\midrule

MIMKD (ours) &&
\textbf{62.91}\Rise{2.94}  & \textbf{62.95}\Rise{2.98} & \textbf{64.32}\Rise{4.35} && 
\textbf{68.77}\Rise{2.68} & \textbf{68.91}\Rise{2.82} & \textbf{69.09}\Rise{3.00}\\
\bottomrule

\end{tabularx}
\end{table*}

\subsection{Computational cost and negative sampling.}
We contextualize the memory and computational overhead of our work with respect to CRD. Our global MI objective has the same footprint as CRD (i.e.~an additional $600\mathrm{MB}$ over standard Resnet18 training for storing negatives). In addition, our feature and local MI objective use projection layers which add an additional~$100\mathrm{MB}$ of GPU memory. As the computation of our JSD-based objective is computationally trivial, we observe negligible reduction in training speed wrt CRD (2.2 epochs/hr v. 2.4 epochs/hr). Note that no additional memory is used for sampling negatives for local and feature information maximization. The $4096$ negatives are only used for global MI as storing 1-D representations is relatively inexpensive.

\subsection{Ablation Study}
In this section we present additional accuracy landscape plots for our extensive ablation study that demonstrates the value of each component of our mutual information maximization objective. We use a ResNet-32x4 as the teacher network and ResNet-8x4 as the student network where the baseline accuracy of the teacher is $79.24 \%$ and that of the student network is $72.44 \%$. The values of the hyper-parameters $\lambda_g$, $\lambda_l$ and $\lambda_f$ --- that control the weight of the global, local and feature mutual information maximization objectives respectively -- were varied between $0$ and $1$ with an increment of $0.25$ while the weight for the cross-entropy loss, $\alpha$ was set to $1$. The following contour plots shows the test accuracy landscape with respect to a pair of hyper-parameters when the third hyper-parameter is set to distinct values. Overall, this demonstrates the value of maximizing region-consistent local and feature-level mutual information between representations in addition to just global information maximization.

%% file: PaperForReview.bbl
\begin{thebibliography}{10}\itemsep=-1pt

\bibitem{ahn2019variational}
Sungsoo Ahn, Shell~Xu Hu, Andreas Damianou, Neil~D Lawrence, and Zhenwen Dai.
\newblock Variational information distillation for knowledge transfer.
\newblock In {\em Proceedings of the IEEE/CVF Conference on Computer Vision and
  Pattern Recognition}, pages 9163--9171, 2019.

\bibitem{belghazi2018mutual}
Mohamed~Ishmael Belghazi, Aristide Baratin, Sai Rajeshwar, Sherjil Ozair,
  Yoshua Bengio, Aaron Courville, and Devon Hjelm.
\newblock Mutual information neural estimation.
\newblock In {\em International Conference on Machine Learning}, pages
  531--540. PMLR, 2018.

\bibitem{brakel2017learning}
Philemon Brakel and Yoshua Bengio.
\newblock Learning independent features with adversarial nets for non-linear
  ica.
\newblock {\em arXiv preprint arXiv:1710.05050}, 2017.

\bibitem{bucilu2006model}
Cristian Buciluǎ, Rich Caruana, and Alexandru Niculescu-Mizil.
\newblock Model compression.
\newblock In {\em Proceedings of the 12th ACM SIGKDD international conference
  on Knowledge discovery and data mining}, pages 535--541, 2006.

\bibitem{chen2017learning}
Guobin Chen, Wongun Choi, Xiang Yu, Tony Han, and Manmohan Chandraker.
\newblock Learning efficient object detection models with knowledge
  distillation.
\newblock In {\em Proceedings of the 31st International Conference on Neural
  Information Processing Systems}, pages 742--751, 2017.

\bibitem{chen2020simple}
Ting Chen, Simon Kornblith, Mohammad Norouzi, and Geoffrey Hinton.
\newblock A simple framework for contrastive learning of visual
  representations.
\newblock In {\em International conference on machine learning}, pages
  1597--1607. PMLR, 2020.

\bibitem{deng2009imagenet}
Jia Deng, Wei Dong, Richard Socher, Li-Jia Li, Kai Li, and Li Fei-Fei.
\newblock Imagenet: A large-scale hierarchical image database.
\newblock In {\em 2009 IEEE conference on computer vision and pattern
  recognition}, pages 248--255. Ieee, 2009.

\bibitem{donsker1983asymptotic}
Monroe~D Donsker and SR~Srinivasa Varadhan.
\newblock Asymptotic evaluation of certain markov process expectations for
  large time. iv.
\newblock {\em Communications on Pure and Applied Mathematics}, 36(2):183--212,
  1983.

\bibitem{gutmann2010noise}
Michael Gutmann and Aapo Hyv{\"a}rinen.
\newblock Noise-contrastive estimation: A new estimation principle for
  unnormalized statistical models.
\newblock In {\em Proceedings of the Thirteenth International Conference on
  Artificial Intelligence and Statistics}, pages 297--304. JMLR Workshop and
  Conference Proceedings, 2010.

\bibitem{he2020momentum}
Kaiming He, Haoqi Fan, Yuxin Wu, Saining Xie, and Ross Girshick.
\newblock Momentum contrast for unsupervised visual representation learning.
\newblock In {\em Proceedings of the IEEE/CVF Conference on Computer Vision and
  Pattern Recognition}, pages 9729--9738, 2020.

\bibitem{he2016deep}
Kaiming He, Xiangyu Zhang, Shaoqing Ren, and Jian Sun.
\newblock Deep residual learning for image recognition.
\newblock In {\em Proceedings of the IEEE conference on computer vision and
  pattern recognition}, pages 770--778, 2016.

\bibitem{hinton2015distilling}
Geoffrey Hinton, Oriol Vinyals, and Jeff Dean.
\newblock Distilling the knowledge in a neural network.
\newblock {\em arXiv preprint arXiv:1503.02531}, 2015.

\bibitem{hjelm2018learning}
R~Devon Hjelm, Alex Fedorov, Samuel Lavoie-Marchildon, Karan Grewal, Phil
  Bachman, Adam Trischler, and Yoshua Bengio.
\newblock Learning deep representations by mutual information estimation and
  maximization.
\newblock {\em arXiv preprint arXiv:1808.06670}, 2018.

\bibitem{huang2018data}
Zehao Huang and Naiyan Wang.
\newblock Data-driven sparse structure selection for deep neural networks.
\newblock In {\em Proceedings of the European conference on computer vision
  (ECCV)}, pages 304--320, 2018.

\bibitem{nowozin2016f}
Sebastian Nowozin, Botond Cseke, and Ryota Tomioka.
\newblock f-gan: Training generative neural samplers using variational
  divergence minimization.
\newblock {\em arXiv preprint arXiv:1606.00709}, 2016.

\bibitem{oord2018representation}
Aaron van~den Oord, Yazhe Li, and Oriol Vinyals.
\newblock Representation learning with contrastive predictive coding.
\newblock {\em arXiv preprint arXiv:1807.03748}, 2018.

\bibitem{paninski2003estimation}
Liam Paninski.
\newblock Estimation of entropy and mutual information.
\newblock {\em Neural computation}, 15(6):1191--1253, 2003.

\bibitem{peng2019correlation}
Baoyun Peng, Xiao Jin, Jiaheng Liu, Dongsheng Li, Yichao Wu, Yu Liu, Shunfeng
  Zhou, and Zhaoning Zhang.
\newblock Correlation congruence for knowledge distillation.
\newblock In {\em Proceedings of the IEEE/CVF International Conference on
  Computer Vision}, pages 5007--5016, 2019.

\bibitem{romero2014fitnets}
Adriana Romero, Nicolas Ballas, Samira~Ebrahimi Kahou, Antoine Chassang, Carlo
  Gatta, and Yoshua Bengio.
\newblock Fitnets: Hints for thin deep nets.
\newblock {\em arXiv preprint arXiv:1412.6550}, 2014.

\bibitem{ruderman2012tighter}
Avraham Ruderman, Mark Reid, Dar{\'\i}o Garc{\'\i}a-Garc{\'\i}a, and James
  Petterson.
\newblock Tighter variational representations of f-divergences via restriction
  to probability measures.
\newblock {\em arXiv preprint arXiv:1206.4664}, 2012.

\bibitem{sandler2018mobilenetv2}
Mark Sandler, Andrew Howard, Menglong Zhu, Andrey Zhmoginov, and Liang-Chieh
  Chen.
\newblock Mobilenetv2: Inverted residuals and linear bottlenecks.
\newblock In {\em Proceedings of the IEEE conference on computer vision and
  pattern recognition}, pages 4510--4520, 2018.

\bibitem{sau2016deep}
Bharat~Bhusan Sau and Vineeth~N Balasubramanian.
\newblock Deep model compression: Distilling knowledge from noisy teachers.
\newblock {\em arXiv preprint arXiv:1610.09650}, 2016.

\bibitem{simonyan2014very}
Karen Simonyan and Andrew Zisserman.
\newblock Very deep convolutional networks for large-scale image recognition.
\newblock {\em arXiv preprint arXiv:1409.1556}, 2014.

\bibitem{tian2019contrastive}
Yonglong Tian, Dilip Krishnan, and Phillip Isola.
\newblock Contrastive representation distillation.
\newblock {\em arXiv preprint arXiv:1910.10699}, 2019.

\bibitem{tung2019similarity}
Frederick Tung and Greg Mori.
\newblock Similarity-preserving knowledge distillation.
\newblock In {\em Proceedings of the IEEE/CVF International Conference on
  Computer Vision}, pages 1365--1374, 2019.

\bibitem{wu2018unsupervised}
Zhirong Wu, Yuanjun Xiong, Stella~X Yu, and Dahua Lin.
\newblock Unsupervised feature learning via non-parametric instance
  discrimination.
\newblock In {\em Proceedings of the IEEE conference on computer vision and
  pattern recognition}, pages 3733--3742, 2018.

\bibitem{yim2017gift}
Junho Yim, Donggyu Joo, Jihoon Bae, and Junmo Kim.
\newblock A gift from knowledge distillation: Fast optimization, network
  minimization and transfer learning.
\newblock In {\em Proceedings of the IEEE Conference on Computer Vision and
  Pattern Recognition}, pages 4133--4141, 2017.

\bibitem{zagoruyko2016paying}
Sergey Zagoruyko and Nikos Komodakis.
\newblock Paying more attention to attention: Improving the performance of
  convolutional neural networks via attention transfer.
\newblock {\em arXiv preprint arXiv:1612.03928}, 2016.

\bibitem{zagoruyko2016wide}
Sergey Zagoruyko and Nikos Komodakis.
\newblock Wide residual networks.
\newblock {\em arXiv preprint arXiv:1605.07146}, 2016.

\bibitem{zeiler2014visualizing}
Matthew~D Zeiler and Rob Fergus.
\newblock Visualizing and understanding convolutional networks.
\newblock In {\em European conference on computer vision}, pages 818--833.
  Springer, 2014.

\bibitem{zhang2019fast}
Feng Zhang, Xiatian Zhu, and Mao Ye.
\newblock Fast human pose estimation.
\newblock In {\em Proceedings of the IEEE/CVF Conference on Computer Vision and
  Pattern Recognition}, pages 3517--3526, 2019.

\bibitem{zhang2018shufflenet}
Xiangyu Zhang, Xinyu Zhou, Mengxiao Lin, and Jian Sun.
\newblock Shufflenet: An extremely efficient convolutional neural network for
  mobile devices.
\newblock In {\em Proceedings of the IEEE conference on computer vision and
  pattern recognition}, pages 6848--6856, 2018.

\end{thebibliography}
